\pdfoutput=1

\documentclass[11pt]{article}

\usepackage{acl}

\usepackage{times}
\usepackage{latexsym}

\usepackage[T1]{fontenc}

\usepackage[utf8]{inputenc}

\usepackage{microtype}

\usepackage{inconsolata}

\usepackage{verbatim}
\usepackage{amssymb}
\usepackage{amsmath}
\usepackage{pifont}
\usepackage{float}
\usepackage{booktabs}
\usepackage{pifont}
\usepackage{tikz}
\usepackage{graphicx}
\title{PRoDeliberation: Parallel Robust Deliberation for End-to-End Spoken Language Understanding}

\author{Trang Le\Thanks{\, Equal contribution} \quad\quad Daniel Lazar\footnotemark[1] \textsuperscript{\, \normalfont 1}\Thanks{\, Work done while at Meta} \quad\quad  Suyoun Kim \quad\quad  Shan Jiang \quad\quad Duc Le \\ \quad\quad \textbf{Adithya Sagar} \quad\quad \textbf{Aleksandr Livshits} \quad\quad \textbf{Ahmed Aly} \quad\quad \textbf{Akshat Shrivastava} \\
Meta, Coldrays\textsuperscript{1} \\ 
\tt{\{trangleminh, dlazar, akshats\}@meta.com}} 

\begin{document}
\maketitle
\begin{abstract}
Spoken Language Understanding (SLU) is a critical component of voice assistants; it consists of converting speech to semantic parses for task execution. Previous works have explored end-to-end models to improve the quality and robustness of SLU models with Deliberation, however these models have remained autoregressive, resulting in higher latencies. In this work we introduce PRoDeliberation, a novel method leveraging a Connectionist Temporal Classification-based decoding strategy as well as a denoising objective to train robust non-autoregressive deliberation models. We show that PRoDeliberation achieves the latency reduction of parallel decoding (2-10x improvement over autoregressive models) while retaining the ability to correct Automatic Speech Recognition (ASR) mistranscriptions of autoregressive deliberation systems. We further show that the design of the denoising training allows PRoDeliberation to overcome the limitations of small ASR devices, and we provide analysis on the necessity of each component of the system. 
\end{abstract}

\section{Introduction}

\begin{figure*}
  \centering
  \includegraphics[width=0.85\linewidth]{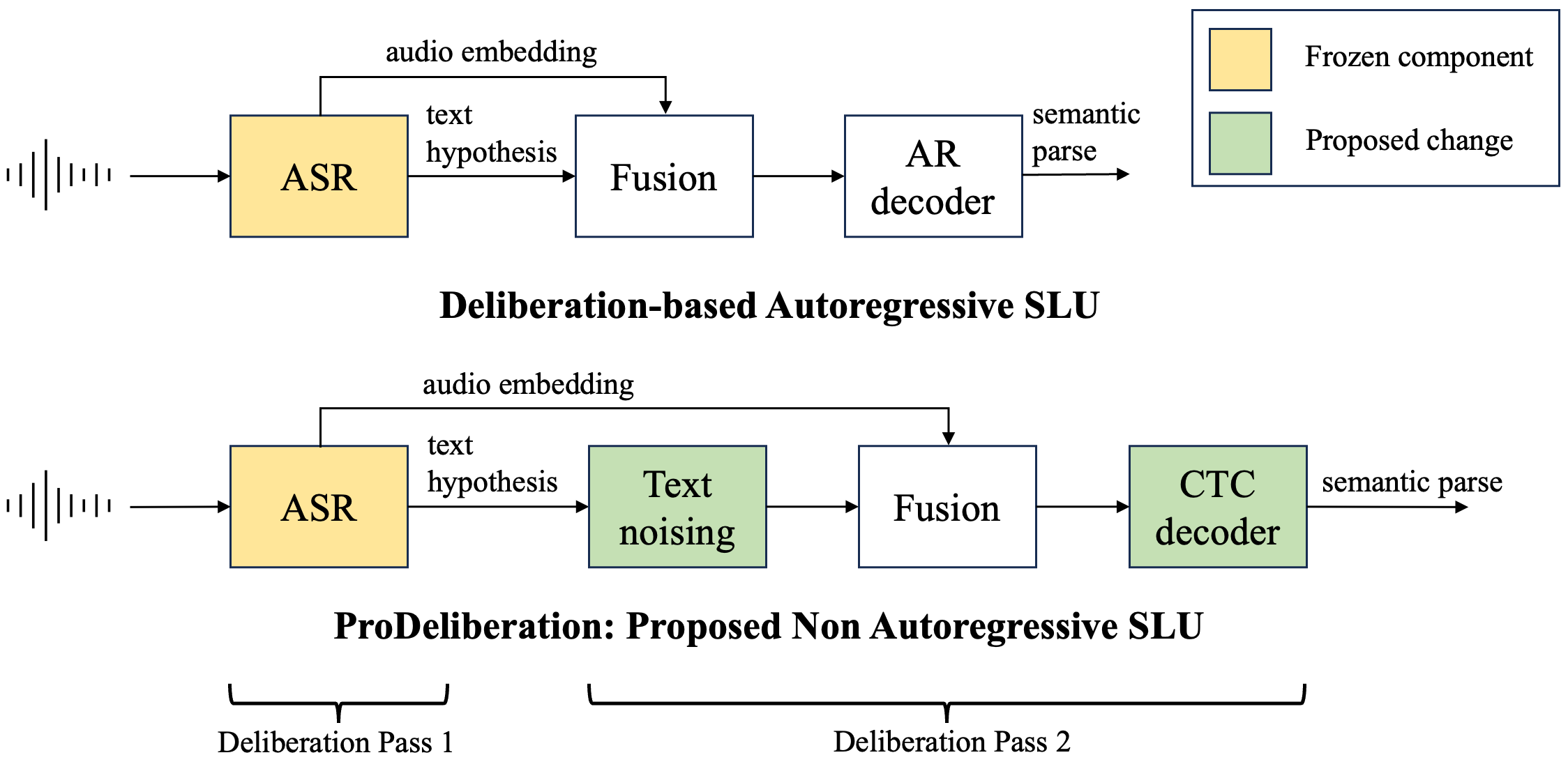}
  \caption{Overview of the proposed method, compared to the previous state of the art for end-to-end Spoken Language Understanding. The proposed method adds two crucial components: a CTC non-autoregressive decoder and a text noising mechanism for training. We label the two passes for the Deliberation architecture (both methods).}
  \label{fig:frontfig}
\end{figure*}

Traditionally, Spoken Language Understanding (SLU) systems are pipelined systems of 1) an Automatic Speech Recognition (ASR) model, which transcribes audio to text, and 2) a semantic parsing model to convert text into structured frames. This results in some inherent limitations, such as cascading errors from ASR mistranscriptions, larger overall systems size from having two models with redundant capabilities, and the inability to use audio cues for semantic parse generation.

End-to-end (E2E) models hold promise of overcoming these limitations by removing the barrier between these two models and allowing audio information to be used in the semantic parse generation. In particular, \textit{deliberation}-based E2E SLU models \cite{le2022deliberation, Hu20deliberation} improve explainability (transcript is produced by first-pass ASR), support of complex compositional structure (via the general decoder), and quality (allows rewriting the ASR transcriptions based on audio information).

However, one downside of the existing deliberation-based E2E SLU models is that they use an autoregressive (AR) decoder for generation, which yields high latency, especially for longer outputs. Previous semantic parsing models can yield very low latency with non-autoregressive (NAR) decoders \cite{shrivastava-etal-2023-retrieve, spanpointernetwork, nar_semantic_parsing, Zhu2020DontPI}, however, these systems use extractive decoders and are therefore unable to correct mistranscriptions from ASR systems. 

In this work we present PRoDeliberation (\textbf{P}arallel \textbf{Ro}bust \textbf{Deliberation}), a novel Connectionist Temporal Classification (CTC) decoder-based E2E SLU model. CTC decoders are commonly used for ASR \cite{Graves2006ConnectionistTC} and have successfully been adapted for machine translation \cite{Libovick2018EndtoEndNN, Saharia2020NonAutoregressiveMT}, but to the best of our knowledge, have never been used for semantic parsing. To further improve the rewrite capabilities of this model, we introduce a \textit{denoising} training regimen inspired by masked language modeling \cite{devlin2019bert, lewis2019bart}, in which we corrupt the transcription text, requiring the model to correct the corruptions by relying on audio. The denoising training applies beyond SLU and can be used with any downstream task which consumes ASR transcriptions.

We study PRoDeliberation and find that PRoDeliberation retains the quality of autoregressive parsers while achieving a \textbf{2-10x} latency reduction across different ASR model sizes. We compare ProDeliberation to Mask Predict-based approaches \cite{ghazvininejad2019maskpredict, nar_semantic_parsing} and confirm that PRoDeliberation overcomes the quality shortcomings of Mask Predict. Finally, we show that the denoising objective further boosts ASR robustness by approximately 0.3\%, thereby exceeding the quality of autoregressive models.

We provide a summary of our approach in Figure~\ref{fig:frontfig}; concretely, our contributions are: 
\begin{enumerate}
    \item We introduce \textbf{PRoDeliberation}, a novel E2E SLU system that leverages CTC and a denoising objective to provide high quality semantic parses at a low latency. 
    \item We study the impact of existing semantic parsing non-autoregressive decoders and find that CTC-based methods are superior for SLU. 
    \item We provide ablations for PRoDeliberation to understand the impact of each component.
\end{enumerate}

\section{Related work}
\label{sec:prior_work}

\subsection{Spoken Language Understanding}

\paragraph{Datasets} Semantic parsing is a critical piece of traditional voice assistants, however text-based semantic parsing benchmarks assume gold text, disregarding errors propagating from speech understanding in the full SLU pipeline. Recently new SLU benchmarks have been introduced such as SLURP \cite{SLURP}, a multidomain SLU benchmark, and STOP \cite{stop2022} a multidomain and compositional SLU benchmark. In this work we study the STOP dataset as it is the largest dataset, includes more speakers, and contains compositionality, making it the most diverse and complex dataset.

\paragraph{Models} \citet{stop2022} propose leveraging fully E2E models pre-trained on the ASR task and fine-tuned on the parsing task to produce a series of models based on HuBERT \cite{HUBERT} and Wav2Vec 2.0 \cite{W2V2}. However such E2E systems remain too large for on-device processing and lack the modularity and interpretablity often required to deploy SLU systems. FANS \cite{Radfar2021FANSFA} explore on-device SLU systems leveraging SLU specific decoders and \citet{BRANCHFORMER} explore distilling E2E models but both lack the interpretability. Neural interfaces \cite{2020NeuralInterface} have been used to retain the modularity of transcription in E2E models, and the Deliberation SLU system \cite{le2022deliberation} further the neural interfaces into the on-device space. In this work we study further optimizing efficiency and robustness of deliberation SLU systems with non-autoregressive decoding and denoising.

\paragraph{Robustness} Recently \citet{MCAT} have explored improving the robustness of Deliberation-based SLU by incorporating confidence signals from ASR models and shown promising results. In this work we take a complimentary direction to improve robustness through denoising training. We focus on denoising as we can scale denoising training arbitrarily influencing the difficulty and robustness training directly.

\subsection{Non-autoregressive Parsing}
The semantic parsing community has shifted towards the seq2seq paradigm \cite{sutskever2014sequence} to improve quality and align with the larger NLP community \cite{lewis2019bart, raffel2020exploring}. With this shift, semantic parsers require low latency due to their real time nature. Non-autoregressive semantic parsing has taken off due to its sub-linear decoding time leading to strong latency improvements. Prior work in this space \cite{nar_semantic_parsing} has shown that non-autoregressive algorithms can lead to lower latency systems at similar accuracy gains. To further improve latency and quality, prior works aim to build an inductive bias regarding semantic parsing systems: \citet{Zhu2020DontPI} rely on insertion transformers to build a semantic parse around the text, \citet{spanpointernetwork} rely on span based prediction to extract slot text, \citet{shrivastava-etal-2023-retrieve} retrieve defined semantic frames and fill in the slots with span prediction.

Our work continues in this direction, critically extending efficient parsing into end-to-end SLU systems where the input is audio and text. The shift away from a gold text input space introduces nuances around robustness and text generation that prior work cannot support due to the focus on extractive decoders. We introduce a CTC loss function \cite{Graves2006ConnectionistTC} for non-autoregressive decoding to reduce the reliance on length prediction, a key bottleneck even in non-end-to-end systems, but even more so for end-to-end SLU.

\section{PRoDeliberation}
In this work we extend Deliberation with two critical components: Firstly, \textbf{CTC decoding}, which utilizes parallel decoding to optimize latency. However, we notice an increased error rate due to parallel decoding. In order to combat this we add the second key component:  \textbf{denoising}, which improve model robustness during training.
\subsection{Deliberation}

\citet{le2022deliberation} introduced the deliberation-based E2E SLU system, shown in the upper part of Figure~\ref{fig:frontfig}. Deliberation was introduced in \citet{Hu20deliberation} as a two-pass ASR system to improve transcription quality. We describe these two passes in detail in the following.

\paragraph{First Pass: Speech Recognition with RNN-T}
The first pass of the system uses the recurrent neural network transducer (RNN-T) \cite{Graves12transduction}, which is the most popular streaming ASR architecture. RNN-T is built on three core components: (1)~\textbf{Acoustic encoder} which consumes the raw audio and produces an audio encoding $emb_{\text{aud}} \in \mathbb{R}^{A \times D}$ (2)~\textbf{Predictor} (internal language model) that consumes the text predicted so far and produces a language encoding $emb_{\text{text}} \in \mathbb{R}^{T \times D}$ and (3)~\textbf{Joiner} which combines these two embeddings $emb_{\text{text}}$ and $emb_{\text{aud}}$ to produce a probability output for the currently predicted token. Each component of RNN-T is streamable, allowing the full model to decode text as the user speaks.

\paragraph{Second Pass: Deliberation Decoder}
The deliberation decoder conducts a second pass over the RNN-T hidden states using the language embedding $emb_{\text{text}}$ and the audio embedding $emb_{\text{aud}}$ that have been cached during the first pass. The deliberation decoder fuses the embedding:
\begin{align*}
    emb_{\text{attn}} &= \texttt{MHA}(emb_{\text{text}}, emb_{\text{aud}}, emb_{\text{aud}}) \\
    emb_{\text{stack}} &= \texttt{Stack}(emb_{\text{text}}, emb_{\text{attn}}) \\
    emb_{\text{fused}} &= \texttt{Linear}(emb_{\text{stack}}) \\ 
    emb_{\text{pool}} &= \texttt{Transformer}(emb_{\text{fused}}) \, ,
\end{align*}
where \texttt{MHA} indicates multi-headed attention. The fused embeddings are then passed to a series of transformer encoder layers to pool the embedding, denoted $emb_{\text{pool}}$. Finally the pooled layers are passed to an autoregressive pointer generator decoder which produces tokens which are either `pointers' to input tokens or newly `generated' tokens \cite{decoupled, see2017get}.

\subsection{Nonautoregressive Decoding with CTC}
\label{sec:ctc}

\begin{figure*}
  \includegraphics[width=\linewidth]{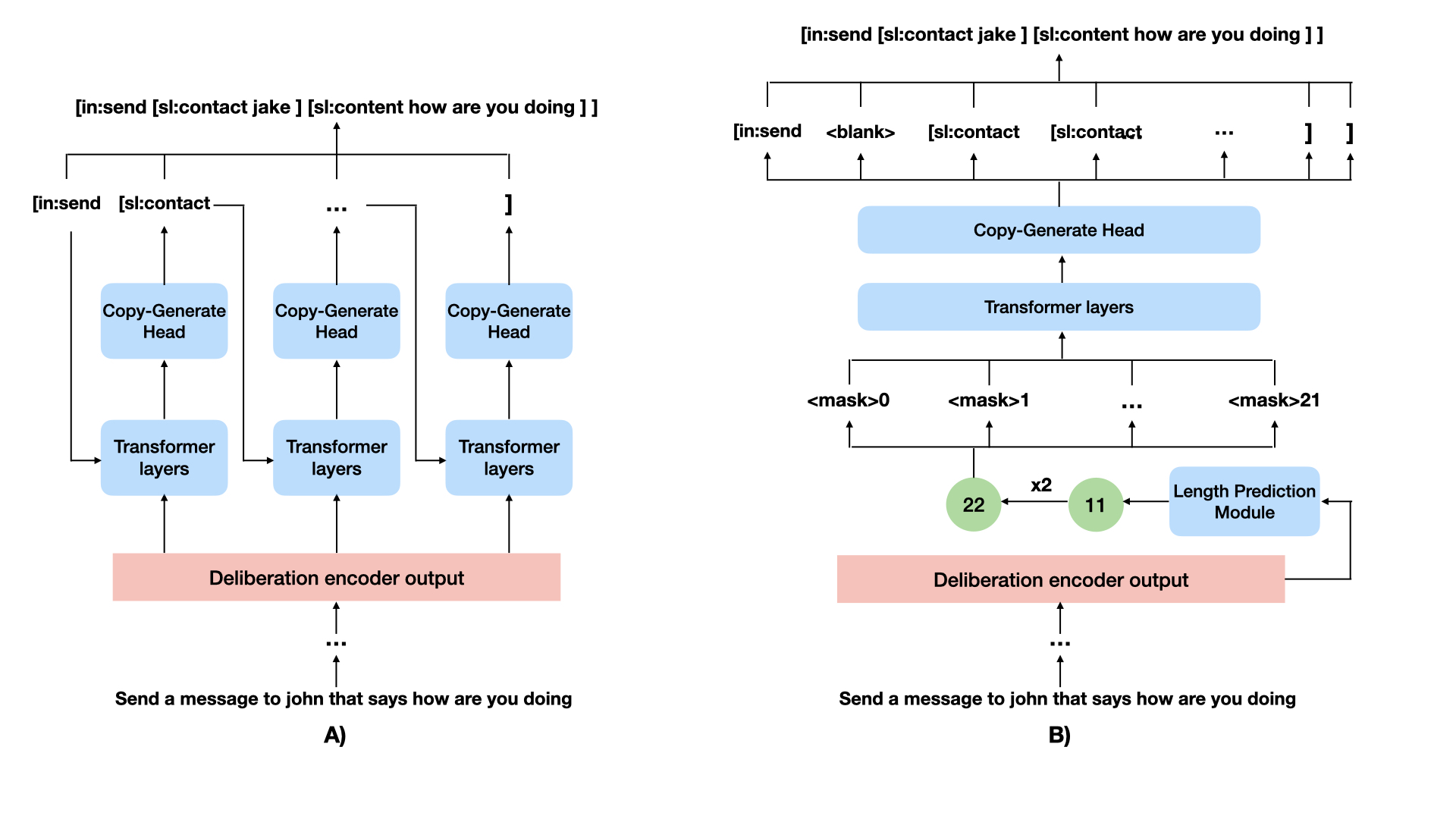}
  \vspace{-40pt}
  \caption{A comparison between (A) Autoregressive Deliberation Decoders and (B) PRoDeliberation architecture. PRoDeliberation extends Deliberation (A) with a fuzzy length prediction module and a CTC based decoder. Both (A) and (B) can correct ASR errors (`john' to `jake' in the figure).}
  \label{fig:ctc_architecture}
\end{figure*}

The latency of autoregressive decoders suffers as sequence length grows due to the linear relationship between latency and sequence length, a result of the conditional dependence of each token output on the preceding tokens. To alleviate this, and following text-based semantic parsing work, we extend deliberation to leverage non-autoregressive decoders~\cite{Zhu2020DontPI, nar_semantic_parsing, spanpointernetwork}. However we find that prior work relies on Mask Predict \cite{ghazvininejad2019maskpredict} which introduces a bottleneck in the form of length prediction of the output sequence. 
In this section we introduce our CTC-based non-autoregressive decoder which overcomes this limitation of length prediction. Our decoder is composed of 2 steps: (1) Length Prediction and (2) Parallel Decoding.

\textbf{Length Prediction} A critical component of 
non-autoregressive networks is length prediction for output sequence ($l$). Prior work implements a length prediction module to determine the output length \cite{nar_semantic_parsing, ghazvininejad2019maskpredict}.

In the extractive setting such as pipelined SLU, where the semantic parse contains either 1) ontology tokens (intent and slot labels) or 2) spans of transcribed text, this does not pose a problem \cite{spanpointernetwork}. However, a major advantage of E2E SLU is the ability to correct ASR errors. This changes the simple extractive function of the decoder into a more complex one, one which may change the length of a span of tokens in a slot from the length of the span predicted by ASR. Because of this, we propose using a CTC decoder which allows for more flexibility in the decoding process than Mask Predict.

Instead of an equality constraint between the predicted length and the final output length, our use of the CTC decoder enforces the constraint that the predicted target length has to be strictly greater than the true target length. 
In semantic parsing the relation between input and output length depends on the semantics and ontology\footnote{An example: "Remind me to call Samantha" -> [IN:CREATE\_REMINDER [SL:TODO call Samantha ] ] vs [IN:CREATE\_REMINDER [SL:TODO [IN:CALL [SL:CONTACT Samantha ] ] ] ] } unlike Machine Translation \cite{Libovick2018EndtoEndNN, Saharia2020NonAutoregressiveMT} or ASR \cite{Graves2006ConnectionistTC} which rely on heuristics based on input length. Accordingly, we introduce a \textbf{fuzzy length predictor}, an extension to the length prediction module that multiplies the output by a fuzzy constant $\alpha>1$ to account for an upper bound. The number of tokens predicted by the decoder, $l$, is then
\begin{align*}
    l = \alpha \times \texttt{LengthModule}(emb_{\text{pool}}) \; .
\end{align*}
This constant helps ensure that even if the length prediction is smaller than the true target length, the decoder is able to output enough tokens for the full semantic parse.

\paragraph{Parallel Decoding with CTC} 
The use of CTC loss creates the flexibility in length prediction by enabling the decoder to produce blank tokens, allowing it to decode any sequence length that is shorter than the fuzzy predicted length.

Our decoder leverages the deliberation encoder's hidden states and length module's [MASK] tokens to decode a semantic parse non-autoregressively:
\begin{align*}
    &y_{1...l} = \texttt{Decoder}([\text{MASK}]_{1...l}, emb_{\text{pool}}) \; ,
\end{align*}
where the decoder is a randomly-initialized Transformer decoder. As a result of CTC loss, the predicted parse $y_1, ..., y_l$ can have repeating tokens and blank tokens. We then apply the CTC alignment procedure \cite{Graves2006ConnectionistTC}, which merges repeating tokens and removes blank tokens to produce the final prediction.

\subsection{Denoising}
\label{sec:denoising}

\begin{figure*}
  \centering
  \includegraphics[width=0.75\linewidth]{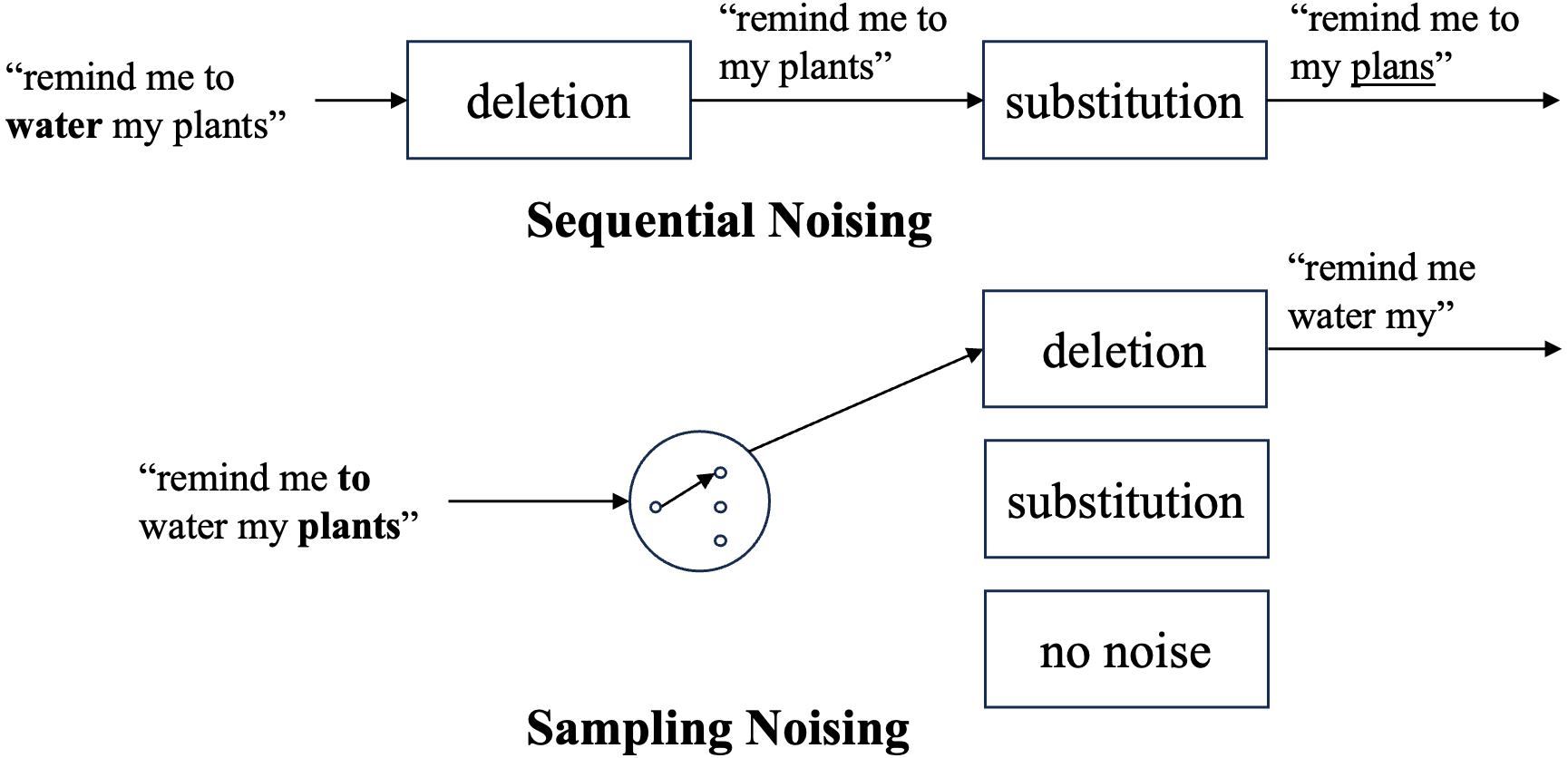}
  \caption{The text noising process. We proposed two noise processes, deletion and substitution noise, as well as two meta-operations, sequential and sampling noising.}
  \label{fig:denoising}
\end{figure*}
While the CTC decoder gives the model the ability to rewrite transcription errors, we wish to specialize training to specifically 
encourage the model to rely on both audio and text signals.
Inspired by the success of Masked Language Modeling \cite{devlin2019bert, lewis2019bart} as well as SpecAugment for ASR training \cite{park2019specaugment}, we add additional text corruptions to the ASR hypotheses to add an implicit \textit{denoising} objective during training. Unlike SpecAugment, the noise impacts the \textit{text} in order to force the model to rely on audio representations in order to correct transcription errors. We use two noising methods, deletion and substitution noise, along with two meta operations: sampling and chaining, summarized in Figure~\ref{fig:denoising}. 

\textbf{Deletion}. Inspired by the BART MLM Objective \cite{lewis2019bart}, we implement a noising function which randomly deletes tokens from the input. Critically, this deletion noise does not insert a MASK token; instead the model must rely on the audio modality to find the corruption and find an appropriate replacement. We choose this approach to mirror the impact of ASR deletions. To implement the noise we employ a binomial distribution where the number of trials is the sequence length, and we introduce a hyper parameter $P$ that controls the probability at each trial to determine the number of tokens ($N$) corrupted from the input. We then uniformly select $N$ tokens from the input and delete them.

\textbf{Substitution}. We also noise input words via word substitution. To find word substitution pairs which emulate true ASR errors we build a dictionary of word confusion pairs from inference of the 10M ASR model over the STOP train set. Similar to deletion noise, we use a binomial distribution to select words for substitution. A word only has substitution applied if it appears in the substitution dictionary. If it does, we sample a replacement word according to the frequency each replacement has appeared for it in the ASR output. 

\textbf{Meta Operations}. With the different noising functions we implemented above, we introduce two meta noising functions that allow us to combine them. 1) Sampling Noise: This mechanism uniformly samples amongst the $R$ different noise types following the parameterization above. 2) Sequential Noise: This mechanism applies $R$ different noise types sequentially, each following the parameterization above. In other words, all samples have each noise type applied in series.

\subsection{Training}

We jointly optimize for training the decoder and length prediction module with a label loss and length loss, respectively. The length loss is negative log likelihood between predicted lengths and target lengths. The label loss is a CTC loss between decoder predictions and target labels. We add label smoothing to both losses to control for over-confident predictions. The overall loss is a weighted sum of these two losses
\begin{align*}
    L = L_{label} + \lambda \times L_{length} \; .
\end{align*}

\section{Experiments}
\label{sec:results}

\subsection{STOP Dataset}
We leverage the STOP (Spoken Task Oriented Parsing) \cite{stop2022} to evaluate our proposed approach. STOP is based on the TOPv2 dataset \cite{chen-2020-topv2}, a compositional multi-domain semantic parsing benchmark that has been well-studied in text parsing. STOP consists of 8 different domains: alarm, messaging, music, navigation, timer, weather, reminder, and event. Each data point contains an audio clip from Amazon Mechanical Turk workers. The  dataset is split into three subsets: 120k training samples, 33k validation samples, and 76k test samples.

\subsection{Baselines}
Our study targets improving Deliberation based SLU systems \cite{le2022deliberation} for on-device streaming use cases. We align our baselines with \citet{le2022deliberation} to study our contributions. 

\paragraph{Base ASR Model} Our studies leverage two on-device-sized ASR models containing 10M and 25M RNNT \cite{Graves12transduction} variants. These models use an acoustic encoder-based conformer encoder \cite{gulati2020conformer, Shi2022conformer} (3L for 10M parameter model and 13L for 25M model), a predictor with a 1-layer LSTM, and a 1-layer Joiner network. We used a 4-stride, 40ms lookahead, 120ms segment size audio input features, and 4000 sentence piece targets \cite{kudo-richardson-2018-sentencepiece}. We employed the same training recipe as proposed originally for E2E deliberation-based SLU \cite{le2022deliberation}. The Word Error Rates (WER) for each split of the STOP datasets, using ASRs of two different sizes (10M and 25M), are presented in Table \ref{tab:asr}.

\paragraph{Autoregressive (AR) Deliberation} Following \citet{le2022deliberation}, we constrain all our model architectures to leverage 5M parameters to enable an iso-size comparison. All models also use the same decoding vocabulary, consisting of 4095 unigram WordPieces plus the vocabulary of ontology tokens built from training data. The 5M AR deliberation model leverages 2 pooling (encoder) layers, 1 decoder layers, with hidden dimension of 224. This AR deliberation model has the best quality and latency performance across various encoder and decoder depths (Table \ref{tab:ar_varying_layers}, Appendix \ref{sec:appendix}).

\paragraph{Mask Predict Deliberation} Our first non-autoregressive architecture extends the AR decoder with the Mask Predict formulation. We introduce a length module after the AR deliberation encoder to predict the output length for non-autoregressive decoding. Iterative refinement is not employed in order to focus on efficient one-shot decoding with minimal latency. Our 5M Mask Predict Deliberation model leverages 4 pooling (encoder) layers, 3 decoder layers, with hidden dimension of 200.

\paragraph{PRoDeliberation} We introduce a fuzzy length prediction and replaces the mask predict loss with a CTC loss to mitigate the length bottleneck. To optimize for latency reduction, iterative refinement is not utilized. Our 5M ProDeliberation model leverages 4 pooling (encoder) layers, 3 decoder layers, with hidden dimension of 200.

\subsection{Training Hyperparameters}
All baseline and proposed models are trained with 8 GPUs of 32GB memory each, using AdamW optimizer and tri-stage learning rate scheduler. All training is done with SpecAugment enabled \cite{park2019specaugment}.

Hyperparameters, including warmup steps, hold steps, decay steps, learning rate, encoder and decoder dropouts, scaling factors in loss, and denoising probabilities, are obtained through hyperparameter sweeps using Bayesian optimization to maximize Exact Match (EM) accuracy on the eval split of STOP dataset. These hyperparameter values can be found in Table \ref{tab:hyperparams}, Appendix \ref{sec:appendix}.

\begin{table}[h]
    \caption{The WER for each split of the STOP datasets, using the frozen RNNT ASR of two different sizes.}
    \centering
    \begin{tabular}{rrrr}
    \toprule
              & \multicolumn{3}{c}{\textbf{WER}} \\
    \textbf{Frozen ASR (RNNT)} & \textbf{train}  & \textbf{valid}  & \textbf{test}  \\
    \midrule
    10M               & 7.68    & 6.99  & 6.54  \\
    25M               & 4.19    & 3.83  & 3.54  \\
    \bottomrule
    \end{tabular}
\label{tab:asr}
\end{table}

\section{Results}

\setlength{\tabcolsep}{4pt}
\begin{figure*}[h!]
  \begin{minipage}[b]{.48\linewidth}
    \centering
    \captionof{table}{Evaluation on the STOP test split.
    PRoDeliberation outperforms Mask Predict, is on par with AR model for 10M ASR, and outperforms for 25M ASR.}
    \begin{tabular}{ccc}
    \toprule
    \textbf{ASR} & \textbf{NLU} & \textbf{EM}  \\
    \midrule
    10M & Deliberation AR \small{\cite{le2022deliberation}}    & 67.90  \\
        & Deliberation AR \small{\cite{MCAT}}  & \textbf{68.37}   \\
        & Mask predict \small{\cite{ghazvininejad2019maskpredict}}    & 67.23   \\
        & ProDeliberation \small{(ours)} & 68.31  \\
    \midrule
    25M & Deliberation AR \small{\cite{le2022deliberation}}   & 73.87  \\
        & Deliberation AR \small{\cite{MCAT}}   & 74.05   \\
        & Mask predict  \small{\cite{ghazvininejad2019maskpredict}}   & 72.17   \\
        & \textbf{ProDeliberation \small{(ours)}} & \textbf{74.27}  \\
    \bottomrule
    \end{tabular}
    \label{tab:ctc_main_results}
  \end{minipage}
  \hfill
  \begin{minipage}[b]{.48\linewidth}
    \centering
      \includegraphics[width=\linewidth]{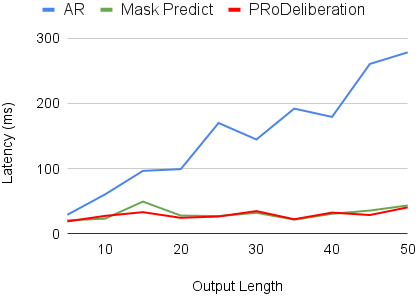}
      \caption{Latency by output length of AR, Mask Predict and ProDeliberation. AR latency increases linearly with output length, while the NAR models (Mask Predict and ProDeliberation) are approximately constant.}
      \label{fig:latency}
  \end{minipage}
\end{figure*}
\setlength{\tabcolsep}{6pt}

\paragraph{Quality Analysis}

On the 10M ASR model, PRoDeliberation achieves better performance than Mask Predict deliberation model and has similar performance to the AR deliberation model. For 25M ASR model, PRoDeliberation outperforms both Mask Predict and AR (Table~\ref{tab:denoising_ablation}). All models are trained on the STOP train split, optimized through hyperparameter sweeps on the STOP validation split, and the best run is evaluated on the STOP test split. The metric used is Exact Match, which is the sample-level accuracy for exactly predicting the target tokens.

\paragraph{Latency Analysis}

To measure the latency of the models, we first use  Torchscript to export the NLU-equivalent module of the baselines and proposed model, each of which consists of fusion layers and an AR decoder or a CTC decoder. We run the Torchscript modules on an X86-64 CPU with 224 GB RAM with various output sequence lengths ranging from 5 to 50 tokens. For each output sequence length, latency is reported by averaging latency measurements of 50 runs.

PRoDeliberation model has significantly lower latency than AR deliberation model; as shown in Figure~\ref{fig:latency}, it is {\textasciitilde}2x faster for short output sequences (<10 tokens) and {\textasciitilde}10x faster for long output sequences (>45 tokens). PRoDeliberation has similar latency to Mask Predict deliberation, which is expected as both models have non-autoregressive decoders. Due to their non-autoregressive decoding nature, their latency is largely unaffected by the output length, while AR deliberation latency scales linearly with the output length.

\section{Analysis}
\label{sec:majhead}

\begin{table*}[h]
    \caption{Denoising ablation study. For each ASR model size, we compare denoising strategies.}
    \centering
    \begin{tabular}{ccccccc}
    \toprule
    Denoise strategy  & \multicolumn{3}{c}{\textbf{10M ASR}} & \multicolumn{3}{c}{\textbf{25M ASR}} \\
     & Subs prob & Del prob & \textbf{EM}    & Subs prob & Del prob   & \textbf{EM}  \\
    \midrule
    samp(del, subs) & 0.088   & 0.003   & \textbf{68.31}   & 0.026    & 0.007 & \textbf{74.27}   \\
    del & -   & 0.012   & 67.8   & -   & 0.024 & 74.07   \\
    subs    & 0.069  & -   & 68.02   & 0.04   & - & 74.02   \\
    no noise    & -   & -    & 67.95   & -  & - & 73.96   \\
    \bottomrule
    \end{tabular}
\label{tab:denoising_ablation}
\end{table*}

\subsection{Architecture Ablation}

\paragraph{Encoder vs Decoder Size}
We study the importance of encoder and decoder depths in ProDeliberation by varying these depths while keeping model size constant (5M parameters) by adjusting the hidden dimension accordingly. Increasing encoder depth up to 4 encoder layers leads to better performance, but more than 4 encoder layers do not further improve the model performance (Table \ref{tab:varying_encoder_layers}). We observe similar trend for decoder depth, where 3 decoder layers results in the best performance, but deeper decoder than 3 layers does not lead to further improvement (Table \ref{tab:varying_decoder_layers}). Overall, while keeping total number of parameters constant, changing encoder and decoder depths does not have a significant impact on  model performance.

\begin{table}[h]
    \caption{Exact Match of ProDeliberation on STOP test set, with varying encoder depths at 5M parameters.}
    \centering
    \begin{tabular}{ccccc}
    \toprule
    \textbf{Enc.} & \textbf{Dec.}   & \textbf{Hidden} & \textbf{Total}   & \textbf{EM}  \\
    \textbf{layers} & \textbf{layers}   & \textbf{dim.} & \textbf{params}   &   \\
    \midrule
    2   & 3 & 224   & 5M    & 66.45  \\
    3   & 3 & 208   & 5M    & 67.84  \\
    4   & 3 & 200   & 5M    & \textbf{68.31}  \\
    5   & 3 & 184   & 5M    & 67.93  \\
    6   & 3 & 176   & 5M    & 68.12  \\
    \bottomrule
    \end{tabular}
\label{tab:varying_encoder_layers}
\end{table}

\begin{table}[h]
    \caption{Exact Match of ProDeliberation on STOP test set, with varying decoder depths at 5M parameters.}
    \centering
    \begin{tabular}{ccccc}
    \toprule
    \textbf{Enc.} & \textbf{Dec.}   & \textbf{Hidden} & \textbf{Total}   & \textbf{EM}  \\
    \textbf{layers} & \textbf{layers}   & \textbf{dim.} & \textbf{params}   &   \\
    \midrule
    4   & 1 & 216   & 5M    & 67.18  \\
    4   & 2 & 208   & 5M    & 67.73  \\
    4   & 3 & 200   & 5M    & \textbf{68.31}  \\
    4   & 4 & 184   & 5M    & 68.05  \\
    4   & 5 & 176   & 5M    & 68.06  \\
    \bottomrule
    \end{tabular}
\label{tab:varying_decoder_layers}
\end{table}

\paragraph{Impact of Target length scale}

As mentioned in Section \ref{sec:ctc}, we scale the predicted output length by a factor $\alpha$ to ensure the sequence length for the CTC loss is greater than or equal to the true output length. Here we investigate how this scale factor $\alpha$ affects the latency and quality of ProDeliberation.

\begin{figure}
  \includegraphics[width=\linewidth]{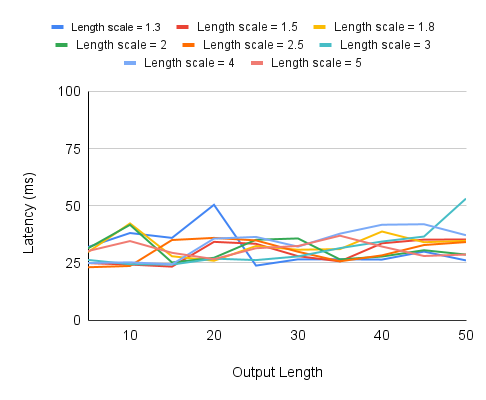}
  \caption{Latency by output length with different target length scale factor $\alpha$.}
  \label{fig:latency_target_length}
\end{figure}

Figure \ref{fig:latency_target_length} shows a minimal impact on latency when we vary target length scale $\alpha$. This is expected as ProDeliberation decodes all output tokens at the same time, so output sequence length does not significantly effect inference latency. 

We observe that $\alpha$ being too small (<1.8) leads to very poor performance, while large $\alpha$ (>2) does not affect performance nearly as much, and $\alpha$ = 2 results in the best performance (Table~\ref{tab:target_length_scale_EM}, Appendix~\ref{sec:appendix}).

\subsection{Denoising ablation}
\label{ssec:denoising_ablation}

We perform an ablation study on the denoising, investigating the relative importance of substitution and deletion noise, as well as the relative performance of the meta noising strategies. The results are summarized in Table~\ref{tab:denoising_ablation}. For each setting, we sweep to find the optimal denoising parameter(s).

Sampling between deletion and substitution noise shows the best performance for both 10M and 25M ASR. The sampling denoise strategy requires higher substitution probability than deletion probability in both 10M and 25M models.
The substitution-only denoise model also performs better than the deletion-only denoise model with 10M ASR and is on-par with the deletion-only denoise model with 25M ASR. 10M ASR requires lower substitution probability but slightly higher deletion probability than 25M ASR. All models with denoising perform better than models with no noise. Moreover, performance without denoising is on par with the autoregressive model; denoising improves performance further.

From the error breakdown, we see that when the ASR model is smaller (and therefore more error prone), a greater share of the improvement from denoising comes from the samples with ASR errors. Moreover, there is greater \textit{overall} improvement with a smaller ASR model. This indicates that the denoising works as expected: by improving robustness to ASR errors.

\setlength{\tabcolsep}{4pt}
\begin{table}
    \caption{Exact Match of E2E models with different ASR model sizes. We break down EM into samples with ASR errors and samples without ASR errors.}
    \centering
    \begin{tabular}{rrrrrrr}
    \toprule
     & \multicolumn{2}{c}{\textbf{10M ASR}} & \multicolumn{2}{c}{\textbf{25M ASR}} \\
    EM & CTC & CTC+D  & CTC & CTC+D  \\
    \midrule
    total  & 67.95  & \textbf{68.31}  & 73.96    & \textbf{74.27}  \\
    no ASR error   & \textbf{85.18}  & 85.14 & 82.95    & \textbf{83.14}  \\
    ASR error    & 25.70  & \textbf{27.00} & 54.36    & \textbf{54.73}  \\
    \bottomrule
    \end{tabular}
\label{tab:asr}
\end{table}
\setlength{\tabcolsep}{6pt}

\section{Conclusion}
\label{sec:conclusion}
We presented a novel SLU model, PRoDeliberation, which uses a CTC decoder and noise augmentation to train a parallelized and robust E2E SLU model. This model achieves much lower latency than the state-of-the-art models while maintaining the ability to correct errors in the ASR transcript. Further, we analyzed the system's performance across two ASR model sizes and provided ablations guiding the decisions made when designing PRoDeliberation. We also show that the denoising training for PRoDeliberation allows it to overcome the limitations of very small ASR models. This represents a significant step forward in building E2E SLU systems that are small enough and have low enough latency to run on edge devices.

\section{Limitations}
\label{sec:limitations}
Even though PRoDeliberation shows significant latency reduction compared to autoregressive deliberation while retaining the model quality, PRoDeliberation still has its own limitations.
\paragraph{Lack of beam search support}
Autoregressive deliberation model can use beam search with larger beam size to improve the model's generation quality, at the cost of added latency. PRoDeliberation, on the other hand, has a non-autoregressive decoder and cannot directly use beam search to further augment the model performance. We can extend PRoDeliberation to allow for beam search in a couple of ways. Similar to \cite{xiao2023survey}, we can use multiple candidate lengths from the length prediction module
and decode according to each in parallel. Another possibility is to look into fusion mechanisms with language models similar to \cite{futami2022non}. However, one thing to note is that in the on-device space which deliberation models target \cite{le2022deliberation}, 
the added latency from larger beam size is rarely tolerated and beam size of 1 is often used.

\paragraph{Data requirements}
This approach requires access to pairs of audio data with semantic parse labels. Though this is a common requirement for end-to-end SLU models, it does pose a more difficult data requirement than cascaded systems, which require audio samples with text labels and text samples with semantic parse labels, the combination of which is easier to generate than the audio data with semantic parse labels.

\bibliography{anthology,custom}

\appendix
\section{Appendix}
\label{sec:appendix}

\label{sec:ar_varying_layers}
\begin{table*}[h]
    \caption{Exact Match and Latency at output length = 10 and output length = 50 of AR Deliberation on STOP eval set, with varying encoder and decoder depths at 5M parameters.}
    \centering
    \begin{tabular}{ccccccc}
    \toprule
    \textbf{Enc.} & \textbf{Dec.}   & \textbf{Hidden} & \textbf{Total}   & \textbf{EM} & \textbf{Latency @} & \textbf{Latency @}\\
    \textbf{layers} & \textbf{layers}   & \textbf{dim.} & \textbf{params}   &   & \textbf{output\_len=10}   & \textbf{output\_len=20}\\
    \midrule
    2   & 1 & 224   & 5M    & 68.37 & 58    & 267  \\
    3   & 1 & 200   & 5M    & 67.85 & 62    & 273  \\
    4   & 1 & 192   & 5M    & 67.86  & 63    & 271 \\
    5   & 1 & 176   & 5M    & 68.11 & 71    & 301  \\
    2   & 2 & 200   & 5M    & 68.27  & 98    & 365 \\
    2   & 3 & 176   & 5M    & 68.39  & 124    & 510 \\
    2   & 4 & 168   & 5M    & 68.02  & 157    & 771 \\
    2   & 5 & 152   & 5M    & 68.37  & 294    & 953 \\

    \bottomrule
    \end{tabular}
\label{tab:ar_varying_layers}
\end{table*}

\begin{table*}[h]
    \caption{Hyperparameter values of PRoDeliberation.}
    \centering
    \begin{tabular}{ccc}
    \toprule
    \textbf{Hyperparameter} & \textbf{10M ASR} & \textbf{25M ASR}\\
    \midrule
    Length loss scaling factor ($\lambda$)   & 0.2504 & 0.1260 \\
    Target length scale ($\alpha$)   & 2 & 2 \\
    Decoder attention dropout   & 0.1784 & 0.4991 \\
    Encoder dropout   & 0.1697 & 0.2646 \\
    Deletion probability   & 0.0026 & 0.0070 \\
    Substitution probability   & 0.0882 & 0.0256 \\
    Learning rate   & 0.00085 & 0.00053\\
    Warm up epochs   & 10 & 10 \\
    Hold epochs   & 40 & 90 \\
    Decay epochs   & 95 & 105 \\

    \bottomrule
    \end{tabular}
\label{tab:hyperparams}
\end{table*}

\begin{table*}[h]
    \caption{Exact Match of PRoDeliberation on STOP test set, with varying target length scale.}
    \centering
    \begin{tabular}{cccccccc}
    \toprule
    \textbf{Target length scale} & 1.3 & 1.5 & 1.8 & 2 & 2.5 & 3 & 4 \\
    \midrule
    \textbf{Exact Match} & 49.50 & 49.95 & 62.44 & \textbf{68.31} & 66.48 & 67.33 & 68.08 \\
    \bottomrule
    \end{tabular}
\label{tab:target_length_scale_EM}
\end{table*}

\end{document}